\documentclass[conference]{IEEEtran}
\usepackage[comma, sort&compress,square,numbers]{natbib}
\usepackage{hyperref}

\ifCLASSINFOpdf
\else
\fi
\usepackage{url}

\usepackage{amsmath}
\usepackage{amssymb}
\usepackage{graphicx}
\usepackage{multirow}
\usepackage{marvosym}
\usepackage{booktabs}
\usepackage{amssymb}
\usepackage{threeparttable}

\newcommand{\corrauthor}{\textsuperscript{\Letter}}

\begin{document}
%
\title{Beyond Source Prompts: Towards Visually Grounded AI-Generated Video Quality Assessment
}

\author{Kehao Zhou\textsuperscript{1},
Yixuan Li\textsuperscript{1}\corrauthor,
Yuxuan Jiang\textsuperscript{1},
Fan Zhang\textsuperscript{1}\\
\textsuperscript{1}\textit{Visual Information Lab, University of Bristol, United Kingdom}\\
\textit{\{or24760, yixuan.li, yuxuan.jiang, fan.zhang\}@bristol.ac.uk}
}

\maketitle

\begingroup
\renewcommand{\thefootnote}{}
\footnotetext{\textsuperscript{\Letter} Corresponding author.\\
This work is funded by the University of Bristol Research Internship programme, and the UKRI MyWorld
Strength in Places Programme (SIPF00006/1). We would like to thank Benoit Vallade from Amazon Prime Video, UK, for his valuable advice and support throughout this work. The authors also
acknowledge the use of resources provided by the Isambard-AI National AI
Research Resource (AIRR). Isambard-AI is operated by the University of
Bristol and is funded by the UK Government’s Department for Science, Innovation
and Technology (DSIT) via UK Research and Innovation and the Science and
Technology Facilities Council [ST/AIRR/I-A-I/1023].}
\endgroup


%


\maketitle


\begin{abstract}
Recent advances in generative artificial intelligence have accelerated video generation technologies. However,  AI-generated videos (AIGVs) remain susceptible to various distortions, highlighting the need for dedicated video quality assessment methods. Content semantics play an important role in AIGV quality assessment, with source prompts typically used as textual references. However, these prompts may not accurately reflect the generated video content, resulting in a discrepancy between the intended textual description and the generated result. To address this limitation, this paper proposes VGCap, an AIGV quality assessment framework guided by visually grounded captions that align better with generated content. Specifically,  we introduce two forms of captions: a global caption obtained by minimally revising the source prompt based on multi-frame visual evidence, and prompt-independent segment captions that describe localized temporal content. Together, these captions provide complementary textual evidence for assessing spatial quality, temporal quality, and text-video alignment. Experimental results on the LGVQ and FETV datasets show that VGCap outperforms representative baselines across all three dimensions. Further ablation studies on caption sources and temporal components confirm the effectiveness of visually grounded captions for each quality dimension, while highlighting the benefits of segment-level features for temporal quality prediction. The source of VGCap will be available at \url{https://github.com/lolo123-123/VGCap}.
\end{abstract}

%
\IEEEpeerreviewmaketitle

\section{Introduction}
\label{sec:intro}

\begin{figure*}[!t]
    \centering
    \includegraphics[width=0.95\textwidth]{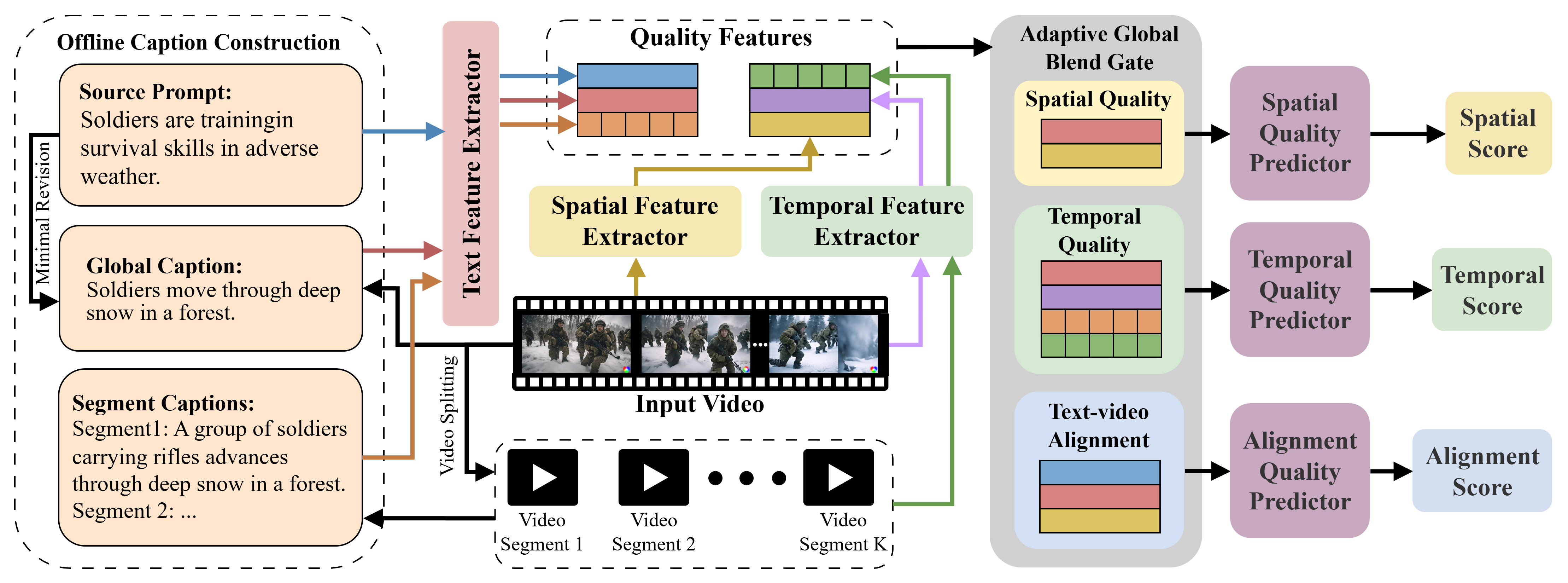}
    \caption{Overall framework of VGCap. The Text Feature Extractor extracts features from the source prompt, global caption, and segment captions. The Spatial Feature Extractor derives frame-level spatial features. The Temporal Feature Extractor produces whole-video and segment-level motion features. The Adaptive Global Blend Gate fuses the branch-specific video and textual features, which are mapped to their corresponding scores by three quality predictors.}
    \vspace{-5pt}
    \label{fig:framework}
\end{figure*}

Recent advances in artificial intelligence-generated content have accelerated text-to-video synthesis~\cite{wu2023tune,khachatryan2023text2video}. However, AI-generated videos (AIGVs) often exhibit generation-specific defects, such as malformed objects, content inconsistency, unnatural motion, and semantic deviations from the source prompt~\cite{huang2024vbench,zhang2025lgvq}. These quality defects need to be captured by assessment models, when the AIGVs are evaluated.

Video quality assessment (VQA) plays an important role in evaluating perceptual video quality and supporting the optimization of video processing systems \cite{bull2021intelligent}. Previous VQA methods have primarily focused on perceptual distortions in natural or user-generated videos. An early method, VSFA, introduced content-aware features and temporal-memory modeling~\cite{li2019quality}. More recent methods, including RAPIQUE and FAST-VQA, integrated statistical and deep features or employed efficient fragment sampling, while DOVER separately modeled technical and aesthetic quality~\cite{tu2021rapique,wu2022fast,wu2023dover}. Although effective for general video distortions, these methods do not explicitly account for generation-specific defects or semantic correspondence with source prompts, limiting their applicability to AIGV quality assessment.

To address these challenges, AIGV quality assessment methods typically consider three complementary dimensions: spatial quality, temporal quality, and text-video alignment~\cite{zhang2025lgvq,wang2025aigvassessor}. Existing methods generally adopt multimodal feature fusion and large vision-language models. Feature-fusion methods include Unify Generated Video Quality Assessment (UGVQ) model, which combines visual, motion, and prompt features~\cite{zhang2025lgvq}, and CRAVE, which jointly models text-video alignment and motion consistency~\cite{sun2025crave}. Large multimodal model-based methods include AIGV-Assessor, which incorporates spatiotemporal and prompt information into large vision-language models (VLMs)~\cite{wang2025aigvassessor}, and VQ-Insight, which adapts VLMs through progressive video-quality learning~\cite{zhang2026vqinsight}.

However, these approaches primarily rely on source prompts as textual references for evaluating semantic correspondence. A source prompt describes the intended video content but may not accurately reflect what is actually generated. Consequently, discrepancies between the source prompt and the generated video may limit the effectiveness of prompt-based features in representing the realized content. This motivates the use of visually grounded captions that directly describe the generated content while preserving the source prompt as a reference for the intended semantics. 



To address these limitations, we propose VGCap, a multidimensional AIGV quality assessment framework that complements the intended semantics of source prompt with visually grounded evidence of the realized video content. VGCap constructs two forms of captions: a global caption obtained by minimally revising the source prompt using multi-frame visual evidence, and prompt-independent segment captions describing localized temporal content. Caption construction is performed offline, decoupling the captioning models from predictor training and inference. Pretrained SigLIP2~\cite{tschannen2025siglip2}, SlowFast~\cite{feichtenhofer2019slowfast}, and BLIP~\cite{li2022blip} encoders extract spatial, motion, and textual features, respectively. Three dedicated branches then fuse the corresponding features to predict spatial quality, temporal quality, and text-video alignment. Experiments show that VGCap outperforms representative baselines across all three quality dimensions. Ablation studies confirm that the global caption improves prediction in all three dimensions and that segment-level motion and caption features further improve temporal quality prediction.

\section{Methodology}
\label{sec:method}

To address the discrepancy between source-prompt and actual video content, VGCap constructs visually grounded captions and integrates them with video features. As shown in Fig.~\ref{fig:framework}, we first describe the offline construction of global and segment captions (Sec.~\ref{ssec:gen_caption}) and then introduce Adaptive Global Blend Gate (AGBG) for fusing video and textual features (Sec.~\ref{ssec:agbg}), followed by the spatial, temporal, and text-video alignment branches (Sec.~\ref{ssec:spatial}--\ref{ssec:alignment}).
\vspace{-4pt}

\subsection{Offline Caption Construction}
\label{ssec:gen_caption}

\subsubsection{Global Caption}
\label{sssec:global_caption}
The global caption refines the source prompt to better describe the realized video content. For videos with at most 32 frames, all frames are retained; otherwise, 32 frames are uniformly sampled. A prompt-verification model decomposes the source prompt into six components: quantity, subject attributes, subject, action or state, object, and scene. Using only the sampled frames, a visual-description model independently generates descriptions for the same components. The prompt-verification model then verifies each prompt component as \textit{supported}, \textit{contradicted}, or \textit{uncertain} based on frame content. Supported and uncertain components are retained, while contradicted components are replaced with the corresponding verified visual descriptions. The components are recomposed into a global caption while preserving the source-prompt structure. 

\subsubsection{Segment Captions}
\label{sssec:segment_captions}
Segment captions describe localized temporal content. Videos with fewer than 16, 16--23, 24--31, and at least 32 frames are evenly divided into one, two, three, and four non-overlapping segments, respectively. Up to eight frames are uniformly sampled from each segment. For each segment, the visual-description model generates an initial caption describing the visible subjects, actions or states, objects, and scene. The prompt-verification model verifies the caption against the corresponding frames. If an inconsistency is detected, the visual-description model revises the caption using the verification feedback, followed by a second verification. This frame-only process keeps the segment captions independent of the source prompt and other captions. A binary mask excludes padded segments during training.

\subsubsection{Caption Construction Details}
The prompt-verification model is implemented using Qwen3-VL-8B~\cite{bai2025qwen3vl}, while the visual-description model uses VideoLLaMA3-7B~\cite{zhang2025videollama3}. Both models are used only during offline caption construction.

\subsection{Adaptive Global Blend Gate}
\label{ssec:agbg}

Video and textual features provide complementary evidence for quality assessment. VGCap therefore uses AGBG to learn their relative contributions during fusion. Before fusion, all input features are projected to a shared feature space $\mathbb{R}^{256}$. Given a video feature $\mathbf{x}$ and a textual feature $\mathbf{y}$, AGBG computes
\begin{equation}
g=\sigma(\gamma),\qquad
\mathbf{h}=(1-g)\mathbf{x}+g\mathbf{y},
\label{eq:agbg}
\end{equation}
where $\gamma$ is a learnable scalar, $\sigma(\cdot)$ is the sigmoid function, $g\in[0,1]$ controls the textual contribution, and $\mathbf{h}$ is the fused feature. Each fusion stage learns an independent gate shared across all samples. The spatial and temporal branches apply this formulation to video-text feature pairs, while the alignment branch extends it to three features using softmax-normalized weights.

\subsection{Spatial Quality Assessment}
\label{ssec:spatial}

Spatial quality assessment benefits from both frame-level visual evidence and semantic context about the realized content. The spatial branch therefore combines spatial video features with the global caption. We first use a pretrained SigLIP2~\cite{tschannen2025siglip2} encoder to extract features from $N$ sampled frames. After projection and positional encoding, a two-layer Transformer produces contextualized frame features $\{\mathbf{h}_i\}_{i=1}^{N}$, where $\mathbf{h}_i$ denotes the feature of the $i$-th frame. We then use three pooling operations to aggregate complementary information across frames, and concatenate and project their outputs to obtain the spatial feature $\mathbf{v}_s$. Specifically, $\bar{\mathbf{h}}$, $\mathbf{h}_a$, and $\mathbf{h}_d$ denote the features obtained by masked mean, attention, and frame-deviation pooling, respectively:
$\bar{\mathbf{h}}=(\sum_i m_i\mathbf{h}_i)/(\sum_i m_i)$,
$\mathbf{h}_a=\sum_i a_i\mathbf{h}_i$, and
$\mathbf{h}_d=\sum_i a_i|\mathbf{h}_i-\bar{\mathbf{h}}|$,
where $m_i\in\{0,1\}$ indicates whether the $i$-th frame is valid and $a_i$ is its normalized attention weight.

To incorporate context from the global caption, we use a pretrained BLIP~\cite{li2022blip} text encoder to extract its feature and project it to $\mathbf{c}$. Following Eq.~\eqref{eq:agbg}, we use AGBG to fuse the features as $\mathbf{z}_s=(1-\alpha_s)\mathbf{v}_s+\alpha_s\mathbf{c}$, where $\alpha_s\in[0,1]$ controls the contribution of the global caption feature and $\mathbf{z}_s\in\mathbb{R}^{256}$ is passed to the spatial quality predictor to produce the spatial score.

\subsection{Temporal Quality Assessment}
\label{ssec:temporal}

Temporal quality depends on both overall motion and localized changes across video segments. We therefore combine whole-video and segment-level motion features with caption features at the corresponding temporal scales. We first use a pretrained SlowFast-R50~\cite{feichtenhofer2019slowfast} encoder to extract a whole-video motion feature from the video and $K$ segment-level motion features from the temporal segments used for segment-caption construction in Sec.~\ref{sssec:segment_captions}. After linear projection, these features are denoted by $\mathbf{v}_t$ and $\{\mathbf{m}_k\}_{k=1}^{K}$, where $K$ is the number of valid segments and $\mathbf{m}_k$ denotes the motion feature of the $k$-th segment. To provide semantic context at the same temporal scales, we use BLIP to extract features from the global caption and segment captions. After linear projection, these features are denoted by $\mathbf{c}$ and $\{\mathbf{s}_k\}_{k=1}^{K}$, respectively, where $\mathbf{s}_k$ denotes the caption feature of the $k$-th segment.

Following Eq.~\eqref{eq:agbg}, we apply AGBG at the whole-video and segment levels as $\mathbf{u}_t=(1-\alpha_t)\mathbf{v}_t+\alpha_t\mathbf{c}$ and $\mathbf{u}_k=(1-\beta_t)\mathbf{m}_k+\beta_t\mathbf{s}_k$, respectively. Here, $\alpha_t,\beta_t\in[0,1]$ control the contributions of the global and segment caption features, while $\mathbf{u}_t$ and $\mathbf{u}_k$ denote the resulting whole-video and $k$-th segment fused features. We then add positional embeddings to $\{\mathbf{u}_k\}_{k=1}^{K}$ and use a temporal Transformer to produce contextualized segment features $\{\mathbf{q}_k\}_{k=1}^{K}$. To aggregate the resulting segment features, we apply three pooling operations, concatenate their outputs, and project them into the same feature space as $\mathbf{u}_t$ to obtain the segment-level temporal feature  $\mathbf{r}_{\mathrm{seg}}$. Specifically, $\bar{\mathbf{q}}$, $\mathbf{q}_a$, and $\mathbf{q}_d$ denote the features obtained by masked mean, attention, and adjacent-segment difference pooling, respectively:
$\bar{\mathbf{q}}=(\sum_{k=1}^{K}\mu_k\mathbf{q}_k)/(\sum_{k=1}^{K}\mu_k)$,
$\mathbf{q}_a=\sum_{k=1}^{K}b_k\mathbf{q}_k$, and
$\mathbf{q}_d=(\sum_{k=2}^{K}\mu_{k-1}\mu_k|\mathbf{q}_k-\mathbf{q}_{k-1}|)/(\sum_{k=2}^{K}\mu_{k-1}\mu_k)$,
where $\mu_k\in\{0,1\}$ indicates whether the $k$-th segment is valid and $b_k$ is its attention weight normalized over valid segments.

Finally, following Eq.~\eqref{eq:agbg}, we combine the whole-video fused feature and segment-level temporal feature using AGBG as $\mathbf{z}_t=(1-\lambda_t)\mathbf{u}_t+\lambda_t\mathbf{r}_{\mathrm{seg}}$, where $\lambda_t\in[0,1]$ controls the segment-level contribution, and $\mathbf{z}_t\in\mathbb{R}^{256}$ is passed to the temporal quality predictor to produce the temporal score.

\subsection{Text-Video Alignment Assessment}
\label{ssec:alignment}

Text-video alignment measures whether the realized visual content matches the intended semantics of the source prompt. Since the global caption describes the realized content, it provides complementary textual evidence for this comparison. We combine $\mathbf{v}_a$, extracted by an independent pathway following $\mathbf{v}_s$ in Sec.~\ref{ssec:spatial}, with BLIP features of the source prompt and global caption, projected as $\mathbf{p}$ and $\mathbf{c}$, respectively.

Extending AGBG in Eq.~\eqref{eq:agbg} to three inputs, we fuse the features as $\mathbf{z}_a=w_V\mathbf{v}_a+w_P\mathbf{p}+w_C\mathbf{c}$, where $w_V$, $w_P$, and $w_C$ are learnable softmax-normalized weights shared across samples. They weight the spatial, source prompt, and global caption features, respectively, and $\mathbf{z}_a\in\mathbb{R}^{256}$ is passed to the alignment quality predictor to produce the alignment score.

\subsection{Training Objectives}
\label{ssec:objectives}

We train the spatial, temporal, and text-video alignment branches independently using their corresponding mean opinion score (MOS) labels, with $b\in\{s,t,a\}$ denoting the three branches. Before loss computation, the prediction $\hat{y}^{(b)}$ and ground-truth MOS $y^{(b)}$ are normalized to $[0,1]$. To optimize score accuracy and ranking consistency, we use $\mathcal{L}^{(b)}=\mathcal{L}_{\mathrm{reg}}^{(b)}+\lambda_r\mathcal{L}_{\mathrm{rank}}^{(b)}$, where $\mathcal{L}^{(b)}$ is the total loss for branch $b$, $\mathcal{L}_{\mathrm{reg}}^{(b)}$ is the Smooth L1 loss~\cite{girshick2015fast}, $\mathcal{L}_{\mathrm{rank}}^{(b)}$ is the pairwise ranking loss~\cite{liu2017rankiqa}, and $\lambda_r$ balances the two losses.

\section{Experiments}
\label{sec:experiments}

\subsection{Experimental Setup}
\label{ssec:setup}

\subsubsection{Evaluation Datasets}
VGCap is evaluated on the Large-scale Generated Video Quality assessment (LGVQ) dataset~\cite{zhang2025lgvq} and the Fine-grained Evaluation of Text-to-Video generation (FETV) benchmark~\cite{liu2023fetv}. LGVQ contains 2,808 videos generated by six models from 468 source prompts and provides MOS labels for spatial quality, temporal quality, and text-video alignment. FETV contains 2,476 videos generated by four models from 619 source prompts; its static-quality, temporal-quality, and overall-alignment scores supervise the corresponding VGCap branches. For both datasets, videos sharing the same source prompt are assigned to the same subset. We construct ten prompt-disjoint splits, each using a 7:1:2 training/validation/test ratio. Performance is evaluated using the Spearman rank-order correlation coefficient (SRCC) and Pearson linear correlation coefficient (PLCC)~\cite{zhang2025lgvq}. Results are averaged over the ten test sets.
\begin{table}[!t]
\centering
\caption{Performance comparison on LGVQ~\cite{zhang2025lgvq} and FETV~\cite{liu2023fetv}. Best and second-best results are bolded and underlined, respectively.}
\label{tab:main_comparison}
\fontsize{4}{4}\selectfont
\setlength{\tabcolsep}{2pt}
\renewcommand{\arraystretch}{1.3}

\resizebox{\columnwidth}{!}{%
\begin{tabular}{@{}lrcc|cc@{}}
\hline
\multirow{2}{*}{Dimension}
& \multirow{2}{*}{Method}
& \multicolumn{2}{c|}{LGVQ~\cite{zhang2025lgvq}}
& \multicolumn{2}{c}{FETV~\cite{liu2023fetv}} \\
\cline{3-6}
& & SRCC & PLCC & SRCC & PLCC \\
\hline

\multirow{6}{*}{Spatial}
& UNIQUE~\cite{zhang2021uncertainty}
& 0.709 & 0.754
& 0.745 & 0.778 \\
& MUSIQ~\cite{ke2021musiq}
& 0.707 & 0.616
& 0.722 & 0.705 \\
& CLIP-IQA~\cite{wang2022exploring}
& 0.672 & 0.593
& 0.698 & 0.629 \\
& LIQE~\cite{zhang2023liqe}
& 0.688 & 0.722
& 0.766 & 0.791 \\
& UGVQ~\cite{zhang2025lgvq}
& \underline{0.764} & \underline{0.793}
& \underline{0.841} & \underline{0.841} \\
& \textbf{VGCap (Ours)}
& \textbf{0.777} & \textbf{0.800}
& \textbf{0.848} & \textbf{0.857} \\
\hline

\multirow{7}{*}{Temporal}
& RAPIQUE~\cite{tu2021rapique}
& 0.659 & 0.609
& 0.833 & 0.854 \\
& VSFA~\cite{li2019quality}
& 0.760 & 0.758
& 0.840 & 0.851 \\
& SimpleVQA~\cite{sun2022simplevqa}
& 0.864 & 0.873
& 0.874 & 0.888 \\
& FAST-VQA~\cite{wu2022fast}
& 0.809 & 0.780
& 0.826 & 0.803 \\
& DOVER~\cite{wu2023dover}
& 0.707 & 0.754
& 0.817 & 0.833 \\
& UGVQ~\cite{zhang2025lgvq}
& \underline{0.894} & \underline{0.910}
& \underline{0.897} & \underline{0.907} \\
& \textbf{VGCap (Ours)}
& \textbf{0.902} & \textbf{0.912}
& \textbf{0.910} & \textbf{0.918} \\
\hline

\multirow{7}{*}{Alignment}
& CLIPScore~\cite{hessel2021clipscore}
& 0.313 & 0.367
& 0.445 & 0.470 \\
& BLIP~\cite{li2022blip}
& 0.499 & 0.537
& 0.520 & 0.572 \\
& ImageReward~\cite{xu2023imagereward}
& 0.503 & 0.546
& 0.607 & 0.654 \\
& PickScore~\cite{kirstain2023pickscore}
& 0.462 & 0.469
& 0.574 & 0.613 \\
& HPSv2~\cite{wu2023humanpreferences}
& 0.376 & 0.401
& 0.539 & 0.573 \\
& UGVQ~\cite{zhang2025lgvq}
& \underline{0.545} & \underline{0.569}
& \underline{0.734} & \underline{0.737} \\
& \textbf{VGCap (Ours)}
& \textbf{0.598} & \textbf{0.616}
& \textbf{0.753} & \textbf{0.789} \\
\hline
\end{tabular}%
}
\end{table}
\subsubsection{Compared Methods}
For a controlled comparison, we focus on encoder-based baselines and exclude assessors built on substantially larger multimodal models. For spatial quality, the baselines include the image quality assessment (IQA) methods UNIQUE~\cite{zhang2021uncertainty}, MUSIQ~\cite{ke2021musiq}, CLIP-IQA~\cite{wang2022exploring}, and LIQE~\cite{zhang2023liqe}, together with UGVQ~\cite{zhang2025lgvq}. For temporal quality, we compare with the VQA methods RAPIQUE~\cite{tu2021rapique}, VSFA~\cite{li2019quality}, SimpleVQA~\cite{sun2022simplevqa}, FAST-VQA~\cite{wu2022fast}, DOVER~\cite{wu2023dover}, and UGVQ. Text-video alignment baselines include CLIPScore~\cite{hessel2021clipscore}, BLIP~\cite{li2022blip}, ImageReward~\cite{xu2023imagereward}, PickScore~\cite{kirstain2023pickscore}, HPSv2~\cite{wu2023humanpreferences}, and UGVQ.

\subsubsection{Implementation Details}
VGCap is trained for 5 epochs using AdamW with a batch size of 4, an initial learning rate of $2\times10^{-4}$, a weight decay of $10^{-4}$, and a cosine learning-rate schedule. In the spatial and alignment branches, SigLIP2 uses the global-caption frame-sampling strategy (Sec.~\ref{sssec:global_caption}), while SigLIP2 and BLIP are frozen in epoch 1 and their final two blocks and layers are fine-tuned thereafter at $2\times10^{-6}$ and $1\times10^{-6}$, respectively. The temporal branch uses fixed, pre-extracted SlowFast and BLIP features. We initialize $\alpha_s=0.33$, $(\alpha_t,\beta_t,\lambda_t)=(0.33,0.45,0.65)$, and $(w_V,w_P,w_C)=(0.55,0.25,0.20)$. The ranking-loss weight $\lambda_r$ and margin are set to $0.3$ and $0.05$, respectively. For each trainable model, the checkpoint with the highest validation SRCC is used for testing. All experiments are conducted on a single NVIDIA GH200 GPU.

All baselines with public implementations are reproduced using their official code and the same data splits as VGCap. For frame-based IQA baselines, we follow the global-caption frame-sampling strategy in Sec.~\ref{sssec:global_caption} and average frame-level predictions to obtain video-level quality scores. Video-based VQA methods follow their original sampling and inference procedures. Alignment measures obtain video-level scores by averaging frame-level image-text alignment or preference scores over the same sampled frames. The UGVQ results are taken from its original paper because its public repository provides the dataset but not the implementation.

\subsection{Performance Comparison}
\label{ssec:comparison}
Table~\ref{tab:main_comparison} compares VGCap with representative image quality, video quality, and text-video alignment assessment methods on LGVQ and FETV. Among the methods, VGCap achieves the best performance across all three quality dimensions in both datasets, with the most pronounced improvement in text-video alignment. These results validate the effectiveness of VGCap's overall design for multidimensional AIGV quality assessment.
\subsection{Ablation Study}
\label{ssec:ablation}
\begin{table}[!t]
\centering
\caption{Ablation study of VGCap's feature components on LGVQ. $\ast$ marks the full configuration for each dimension; best and second-best results are bolded and underlined, respectively.}
\label{tab:feature_ablation}
\fontsize{9}{8}\selectfont
\setlength{\tabcolsep}{4pt}
\renewcommand{\arraystretch}{1.1}
\begin{tabular}{@{}r@{\;}c l l rr@{}}
\toprule
\multicolumn{2}{c}{No.}
& Dimension
& Configuration
& \multicolumn{1}{c}{SRCC}
& \multicolumn{1}{c}{PLCC} \\
\midrule

1 & &
\multirow{3}{*}{Spatial}
& $\mathbf{v}_s$
& 0.761 & 0.782 \\

2 & &
& $\mathbf{v}_s,\mathbf{p}$
& \underline{0.772} & \underline{0.790} \\

3 & $\ast$
& $\vphantom{\mathbf{v}_s}$
& $\mathbf{v}_s,\mathbf{c}$
& \textbf{0.777} & \textbf{0.800} \\

\midrule

4 & &
\multirow{5}{*}{Temporal}
& $\mathbf{v}_t$
& 0.879 & 0.891 \\

5 & &
& $\mathbf{v}_t,\mathbf{p}$
& 0.890 & 0.898 \\

6 & &
& $\mathbf{v}_t,\mathbf{c}$
& 0.894 & 0.905 \\

7 & &
& $\mathbf{v}_t,\mathbf{c},\{\mathbf{m}_k\}_{k=1}^{K}$
& \underline{0.898} & \underline{0.908} \\

8 & $\ast$
&
& $\mathbf{v}_t,\mathbf{c},
   \{\mathbf{m}_k,\mathbf{s}_k\}_{k=1}^{K}$
& \textbf{0.902} & \textbf{0.912} \\

\midrule

9 & &
\multirow{3}{*}{Alignment}
& $\mathbf{v}_a$
& 0.563 & 0.572 \\

10 & &
& $\mathbf{v}_a,\mathbf{p}$
& \underline{0.568} & \underline{0.588} \\

11 & $\ast$
&
& $\mathbf{v}_a,\mathbf{p},\mathbf{c}$
& \textbf{0.598} & \textbf{0.616} \\

\bottomrule
\end{tabular}
\end{table}
Table~\ref{tab:feature_ablation} evaluates the global caption and segment-level features on LGVQ. Replacing the source prompt with the global caption improves both spatial and whole-video temporal assessment. Adding segment-level motion and caption features progressively improves temporal performance. For text-video alignment, combining the spatial feature with both the source prompt and global caption achieves the best result. Overall, the global caption benefits all three dimensions, while segment-level features further improve temporal quality prediction.

\section{Conclusion}
\label{sec:conclusion}

This paper presented VGCap, a multidimensional AIGV quality assessment framework guided by visually grounded captions. VGCap constructs a global caption from multi-frame visual evidence and prompt-independent segment captions for localized temporal content, to obtain better content awareness. Experiments on LGVQ and FETV show strong performance across all three quality dimensions, with the largest gains in text-video alignment. Ablations confirm that the global caption improves all three dimensions and that segment-level features further enhance temporal quality prediction. These results show that visually grounded captions help assess generated videos based on their actual content. A current limitation is that offline caption construction requires additional VLM inference, increasing computational cost. Future work will explore automated caption verification and more efficient integration of caption construction with quality assessment.
\clearpage

\small
\bibliographystyle{IEEEtran}
\bibliography{refs}

\end{document}